# Automated speech-based screening of depression using deep convolutional neural networks

Karol Chlasta[a,b]*, Krzysztof Wołk[a], Izabela Krejtz[b]

[a]*Polish-Japanese Academy of Information Technology, Koszykowa 86, 02-008 Warsaw, Poland*
[b]*SWPS University of Social Sciences and Humanities, Chodakowska 19/31, 03-815 Warsaw, Poland*

**Abstract**

Early detection and treatment of depression is essential in promoting remission, preventing relapse, and reducing the emotional burden of the disease. Current diagnoses are primarily subjective, inconsistent across professionals, and expensive for individuals who may be in urgent need of help. This paper proposes a novel approach to automated depression detection in speech using convolutional neural network (CNN) and multipart interactive training. The model was tested using 2568 voice samples obtained from 77 non-depressed and 30 depressed individuals. In experiment conducted, data were applied to residual CNNs in the form of spectrograms—images auto-generated from audio samples. The experimental results obtained using different ResNet architectures gave a promising baseline accuracy reaching 77%.

* Corresponding author. Tel.: +48-572-885-704; fax: +48-22-584-4501.
  *E-mail address:* karol.chlasta@pja.edu.pl

## 1. Introduction

Globally, more than 300 million people of all ages suffer from depression[†], a common mental disorder. Symptoms include sadness, loss of interest in activities, changes in appetite and sleep patterns, loss of energy, difficulty thinking coherently, and increased possibly of thoughts of suicide [1]. People are often passive in reaching out to clinical psychologists or psychiatrists to get support with their mental health issues. According to the World Health Organization (WHO), fewer than half of those affected in worldwide (in many countries, fewer than 10%) receive treatment against depression. The other fact noted by WHO is that depression is on the rise, with an overall increase of more than 18% in depression cases between 2005 and 2015[‡]. WHO states that, although there are known[§], effective treatments for depression, one of the main barriers to effective care is its inaccurate assessment. Therefore, the development of an easily accessible tool enabling valid and reliable pre-screening of depressive symptomatology is essential. Such a solution should be based not only on a self-assessment questionnaire, but also on an mhealth solution using new technologies, as WHO urged its members during the Seventy-First World Health Assembly in 2018 [2]. Our method could relatively easily be implemented in a mobile phone application or a wearable device. Such an easily accessible system could limit the probability of inaccurate assessment, and lower social stigma, as per a recent WHO recommendation.

Studies have shown a link between depression and human behavior. One of depression's characteristics is a decline of motor coordination (also laryngeal control) that is caused by a change in the brain's basal ganglia. Quatieri and Malyska [3] showed that voice quality contains information about the mental state of a person, and that vocal source features can be used as biomarkers of depression severity. They suggested that commonly used voice quality measures in depression detection could be borrowed from signal-processing methodologies, such as jitter, shimmer, the small cycle-to-cycle variations in glottal pulse amplitude in voiced regions, harmonic-to-noise ratio, and the ratio of harmonics to inharmonic components. These features were related to vocal fold vibration, which were influenced by vocal fold tension and subglottal pressure. The correlations of laryngeal biomarkers with psychomotor retardation assessment resulted in improved understanding of the neurophysiological basis for changes in voice quality with depression and human speech degradation. Several other researchers have discussed how common paralinguistic speech characteristics are affected by depression and the application of this information in classification and prediction systems [4]. These links have a promising potential to be quantified by automatic depression detection (ADD) algorithms.

Scientific investigation of depression was a focal point of the 2017 AVEC "Real-life Depression and Affect Recognition Workshop and Challenge" organized by Association for Computing Machinery [5]. At this conference, Yang et al. [6] proposed a hybrid multi-modal depression recognition and classification framework comprising three sections: the prediction of patient health questionnaire depression scale (PHQ-8) [7] score from audio and video features; the classification of depression/non-depression from text information; and the multimodal regression for final depression prediction. This approach obtained better performance than the baseline results in estimations of PHQ-8 score as measured by the mean absolute error (MAE) [8] and the root-mean-square error (RMSE), with MAE of 4.359, and RMSE of 5.400. The other results were satisfying as well. Paragraph vectors and support vector machine (SVM) [9] models for text-based classification gave accuracies of 84.21% for female subjects and 81.25% for male subjects. Yang et al. [10] proposed a multi-modal fusion framework composed of deep convolutional neural network (CNN) models [11]. Their framework considered audio, video, and text streams, and again, it obtained better performance than the baseline in predicting PHQ-8 scores from audio, video, and text data (MAE 3.980, RMSE 4.653). Multimodal approaches based on different neural network architectures appear to be more effective than those based on a single modality, and they are an interesting area of further research.

---

[†] World Health Organization, Depression Key Facts https://www.who.int/news-room/fact-sheets/detail/depression
[‡] World Health Organization, "Depression let's talk" says WHO, as depression tops list of causes of ill health
https://www.who.int/news-room/detail/30-03-2017--depression-let-s-talk-says-who-as-depression-tops-list-of-causes-of-ill-health
[‡] OpenFace: an open source facial behavior analysis toolkit http://ieeexplore.ieee.org/abstract/document/7477553/

Another set of interesting results was presented during 2018 Interspeech Conference[**]. Specifically, Afshan et al. [12] focused on the effectiveness of voice quality features in detecting depression. They proposed the use of voice quality features in combination with Mel-frequency cepstral coefficients (MFCCs) [13], which showed improved performance in detecting depression. As a result, they were able to achieve an accuracy of 77% even when the test utterances were as short as 10 s. The accuracy was as high as 95% when the test utterances were 1.8 minutes long.

Hanai et al. [14] presented a depression detection model based on sequences of audio and text transcriptions. They evaluated a regularized logistic regression model (with and without conditioning on the type of questions asked), and a long short-term memory (LSTM) model [28] (using the sequences of responses, and without knowledge of the type of questions that prompted the responses). They also evaluated a multi-modal LSTM model that combined the audio and text features. The interesting findings they announced were that context-free modelling of the interviews based on text features performed better than audio features when classifying for a binary outcome (depressed vs. non-depressed). On the other hand, audio features were more accurate in determining the multi-class depression score (MAE 5.01 vs. 7.02). When weighting the model according to the questions asked, audio features performed better than text features (F1 of 0.67 vs. 0.44) with perfect rates of precision (1.00). The overall performance of audio improved when conditioning on the question being asked (F1 0.67 vs. 0.50). According to these researchers the multi-modal model yielded the best performance (F1 0.77 and recall 0.83). Sequence models also displayed the best multi-class classification performance.

This paper proposes a novel method for automated speech-based screening of depression using deep convolutional neural networks. We present comprehensive experiments on distress analysis interview corpus (DAIC) [15] to show the potential of our classification method and evaluate the results obtained.

## 2. Data

### 2.1. Data source and data description

The DAIC database contains clinical interviews designed to support the diagnosis of psychological distress conditions such as anxiety, depression, and post-traumatic stress [15]. These interviews were collected as part of a larger effort to evaluate a computer agent that interviews people with mental illness [16]. The data were published free of charge for scientific use by the University of Southern California[††]. The archive contains 92 GB of data stored as a package of 189 folders compressed in zip format. They contain 300 to 492 session recordings. Each file in the archive represents a single session, and it contains a text transcription of the recording, participant audio files, and facial features. The audio files were recorded with a head mounted microphone (Sennheiser HSP 4-EW-3) at 16 kHz. Interviews were conducted by an animated virtual interviewer called Ellie, controlled by a human interviewer in another room. Figure 1 presents Ellie, the virtual interviewer.

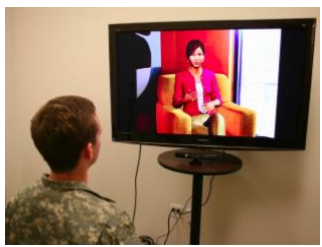

Fig. 1. Ellie, the virtual interviewer [11].

---

[**] Website of Interspeech 2018 Conference https://interspeech2018.org/
[††] Website of DAIC-WOZ Database http://dcapswoz.ict.usc.edu/

The length of each session ranges between 7–33 min (with an average of 16 mins). The level of depression was labelled with a single value per recording using a standardized self-assessed subjective depression scale, PHQ-8 [7]. The authors decided to use the same dataset to allow comparability of results with Depression Sub-Challenge of AVEC'17. We decided to use all the fully annotated (108) audio files of an official AVEC'17 train split, and we grouped the files into two folders by their PHQ-8 score. A score of 10, or more indicated a "depressed" subject, whereas a lower score indicated a "non-depressed" subject.

*2.2. Data pre-processing*

The audio files were cut programmatically using Python‡‡ and Sox§§ into 15 second samples starting from the 60th second of each recording to capture speech samples, rather than background noise. The length of each recording was normalized to 15 s and it turned out to be satisfactory. We wanted to capture a few (2–3) sentences, so that our method could identify key characteristics of speech in the training data, and later match them in the test data. This gave us voice samples of each person that we used in spectrogram generation, succeeded by CNN training and validation. We refer to this original dataset of 107 speech samples as Dataset A. Another dataset was created to check how the method would work on a much larger dataset. The extended Dataset B was created by extracting each 15 s of speech starting from the 60th second of each recording, until the 7$^{th}$ min (the shortest session in DAIC is 7 min). This approach produced 2568 speech samples, 720 of each were of 30 depressive subjects. The training set of the AVEC'17 consists of an imbalanced number of depressed and not-depressed samples, as shown in Table 1. We recognized that such an imbalanced dataset may decrease the recognition performance and cause overfitting. Thus, in the first step we re-sampled the dataset to obtain larger scale data.

Table 1. Number of speech samples in Dataset A (normalized) and Dataset B (extended).

| Dataset A | | Dataset B | |
| --- | --- | --- | --- |
| Category | Samples | Category | Samples |
| Depressed | 30 | Depressed | 720 |
| Non-depressed | 77 | Non-depressed | 1848 |

*2.3. Constructing dataset for machine learning*

Uncompressed audio in wave audio format (WAV)*** is stored as a sequence of numbers that indicate the amplitude of the recorded sound pressure at each point in time. In the WAV standard, these numbers are packed into a byte-string. The interpretation of this byte-string is based on two main factors. The first is sampling rate (usually given in Hertz†††), which indicates how many samples‡‡‡ each second of our audio data consists of. The second is a sample width, which indicates how many bits comprise a single sample (a bit rate§§§). These parameters, along with other parameters (e.g. number of channels: mono or stereo) are stored in the header of each WAV file. Frames normally hold an entire byte-string representing all the audio samples in a sound file. Our task required unpacking of a byte-string for each file into an array of numbers that could be analyzed further. We validated that DAIC-WOZ audio files were indeed recorded at 16 kHz with a 2-byte sample width. Based on the information gathered, we generated wavelet and spectrogram**** for each file. A sample wavelet is presented in Fig. 2.

---

‡‡ Website of Python programming language https://www.python.org/
§§ Website of Sox, a cross-platform command line utility that can convert various formats of computer audio files http://sox.sourceforge.net/
*** WAV (wave form audio format) is a subset of Microsoft's RIFF specification for the storage of multimedia files http://www-mmsp.ece.mcgill.ca/Documents/AudioFormats/WAVE/WAVE.html
††† Hertz, unit of frequency https://www.britannica.com/science/hertz
‡‡‡ Audio sampling rate http://www.digitizationguidelines.gov/term.php?term=samplingrateaudio
§§§ Definition of bit rate from Merriam-Webster dictionary https://www.merriam-webster.com/dictionary/bit%20rate

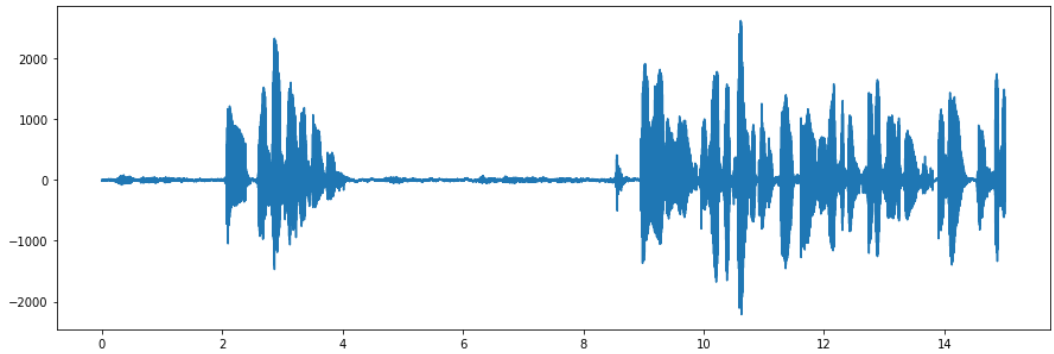

Fig. 2. Sample wavelet representing session 348. First 15 seconds of the 2$^{nd}$ minute of the recording; interviewee PHQ8 score of 20.

Subsequently, the dataset was downsampled with a low pass filter [17]. The filter was applied to prevent aliasing [18] and improve the generalization of the dataset [17]. Figure 3 presents a view of a downsampled file, alongside a spectrogram generated from the same file (in grey).

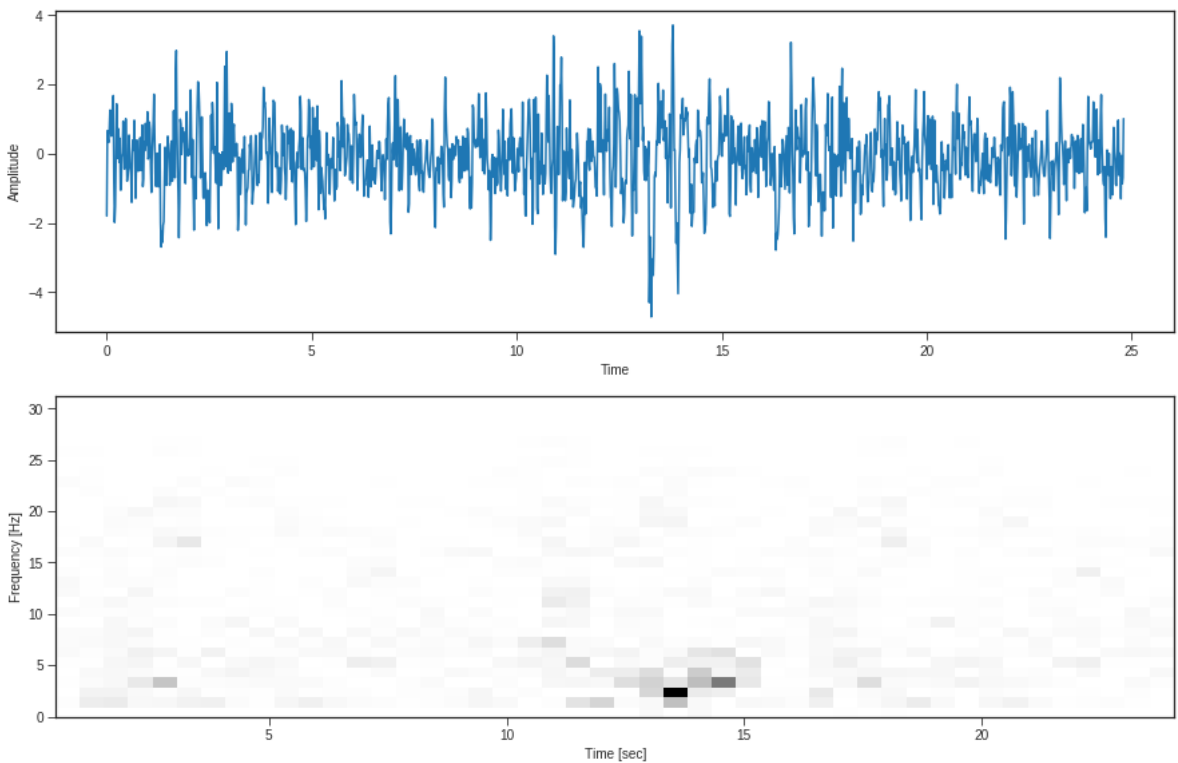

Fig. 3. Example of a downsampled waveform along with a spectrogram from a "depressed" class.

---

**** Definition of spectrogram from Merriam-Webster dictionary https://www.merriam-webster.com/dictionary/spectrogram

The spectrogram of each session file was resized to fit 224 × 224 px size and saved in color. Exemplary images are shown in Fig. 4.

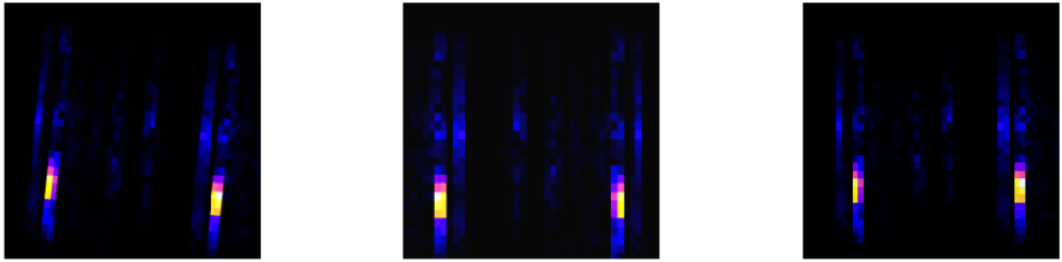

Fig. 4. Example of pre-processed spectrogram with enabled data augmentation, as used in training.

This image format was used with the CNN in training and prediction, as required by pre-trained models [19]. Experiments with resolutions of 512 × 512 px, 1024 × 1024 px, and higher were conducted as well. Finally, the data were split into training and test sets by randomly assigning 75% of the items to the training set, and 25% to the test set.

## 3. Method

In this study, we used several pre-trained CNN architectures that used a residual learning framework (ResNet-18, 34, 50, 101, 152) [20]. As our system was built on a previously trained model, we had to perform fine-tuning on the available ResNet architectures to train a state-of-the-art image classifier [21].

Fig. 5 presents the general schema of the fine-tuning method performed on our pre-trained CNNs. We applied the approach where the last layer of the model is replaced by a layer of dimensions appropriate for the dataset. In our case, layer FC8 was replaced by a new layer with the number of outputs equal to the number of two possible prediction classes (Depressed/Non-depressed), where the frozen layers were the pre-trained ResNet architectures.

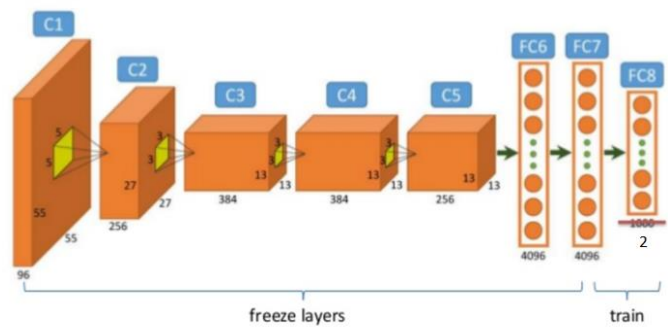

Fig. 5. Schema of our CNN during fine-tuning. Our layer FC8 has two prediction classes[††††]

We fitted the existing model to our data by finding two hyperparameters:
- Learning rate (LR): we used the parameter value of 0.01 or 0.001 as a starting point (See Fig. 6),
- Number of epochs (EP): single passages through the entire training set (See Fig. 7).

---

[††††] Slide 17 of Practical Deep Learning by André Karpištšenko https://www.slideshare.net/AndrKarpitenko/practical-deep-learning

Hyperparameters express high-level properties of the model that cannot be learnt in the normal learning process, but by adjusting the model to the dataset. We performed this adjustment with a stochastic gradient descent with restart (SGDR) [22]. Our system uses multipart interactive training—a training procedure in which the process is interrupted by a human decision at key stages. Finally, we selected the model that provided the best classification accuracy. Our algorithm can be described in nine steps:

1. Enable the data augmentation and precomputed activation cache for CNN training.
2. Find the highest learning rate value, where the loss function's value is still decreasing.
3. Freeze all the CNN layers, except the last one.
4. Train the last layer for 1–2 epochs with precomputed activations.
5. Train the last layer for 2–3 epochs with data augmentation and the *cycle_len* parameter set to one.
6. Unfreeze the frozen layers.
7. Set the learning rate lower by a factor of 3–10 for the previous layer than for the next layer.
8. Find again the highest value of the learning rate, where the loss function value is still decreasing.
9. Train the whole CNN with the *cycle_mult* parameter set to two until network overfitting occurs.

There are four aspects of the method that we would like to comment on in more detail. The first is finding a learning rate, which is relevant for steps 2 and 8 of the algorithm. It is performed by analyzing a loss function, such as that depicted in Fig. 6. It can be seen that the learning rate still improves after lr = 1e-3 (0.001); therefore, we selected a parameter value that minimized our validation loss. Further, in our case of training the ResNet-50 network, we selected a learning rate of 0.005.

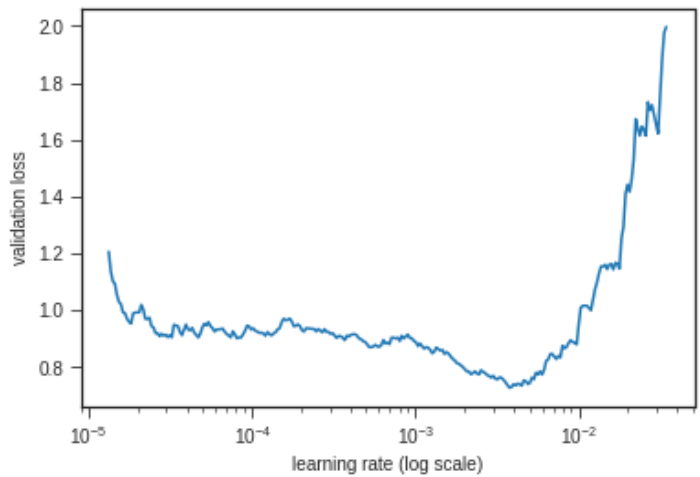

Fig. 6. Learning loss function for ResNet-50 trained on spectrograms of 512x512 px.

The second aspect is a technique that we used in steps 4, 5, and 9. It is called cyclic stochastic gradient descent with restart (SGDR) [22], and it is a variant of annealing the learning rate, which gradually reduces the learning rate, as we progress through the training process. This is helpful because when approaching optimal model weights, it is essential to take smaller steps [23]. This technique relates to setting the values of *cycle_len* and *cycle_mult* parameters that are used in training. The *cycle_len* parameter describes the number of epochs between restarts, which is at points when our system increases the speed of learning. As a result, we force our model to move to another part of the weighing space. The *cycle_mult* parameter indicates a multiplier of the length of the next cycle (by a given value, in our case: 2), applied as soon as the previous cycle is finished. Fig. 7 illustrates an example when the learning rates finding process is restarted (a) once and (b) three times.

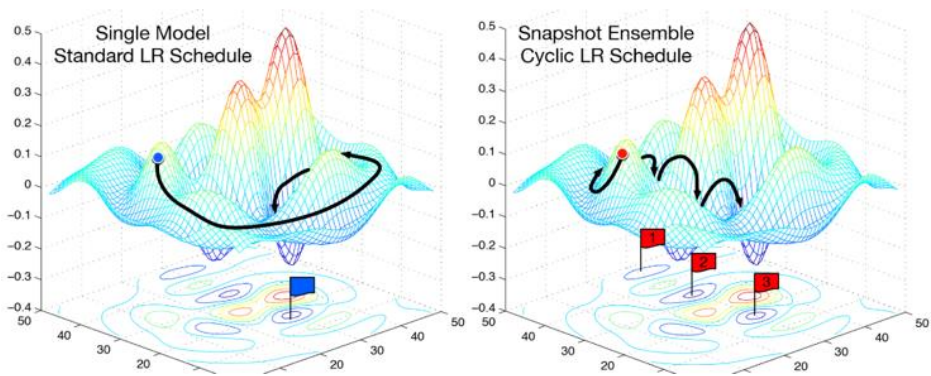

Fig. 7. Cyclic stochastic gradient descent with restart (SGDR): (a) cycle_len = 1; (b) cycle_len = 3 [26].

The third aspect is fine-tuning and learning rate annealing, which is used in steps 7, 8, and 9. It is used when all layers of our neural network are unlocked. Having a well-trained final layer in steps 4 and 5, we refine the remaining layers. The remaining layers had already been trained to recognize images, so the approach does not destroy the accurately pre-adjusted scales of our neural network. At that point we set a 3–10x smaller learning step for the previous layer, than for the next layer. In this model, earlier layers have more general-purpose functions, and they are fine-tuned to the new datasets. As a result, different learning rate levels are used for different layers: the first few layers were trained at level 1e-4 (0.0001), the middle layers at 1e-3 (0.001), and the final layers (see FC layers in Fig. 5) lowered to 1e-2 (0.01).

## 4. Experiments and results

We performed all our experiments using Google Collaboratory platform, which enabled us to use NVidia K100 graphic cards for computations. It also uses Jupyter Notebook standard, facilitating the exchange of code and results with other researchers. Our Python code and results can be accessed easily by cloning a public Jupyter notebook[‡‡‡‡]. We used PyTorch and fast.ai[§§§§] for neural network training; in our opinion, this accelerated the development process and facilitated application of best practice (e.g. this approach allowed us to evaluate multiple pre-trained ResNet architectures by changing a single parameter value). Our classification results were evaluated using four classification assessment metrics [24]: Accuracy, F1 Score (F-measure), Precision, and Recall (Sensitivity).

The experiments included the data and CNN preparation, as well as fine-tuning procedure required to obtain the state-of-the-art results for our automated speech-based depression screening system. Apart from evaluating five CNN architectures (ResNet-18, 34, 50, 101, 152), we also checked if generating higher resolution spectrograms would impact the classification results. We confirmed that generating a set of larger input spectrograms (e.g. 1024x1024) would not significantly improve the results. We also performed Test Time Augmentation (TTA) [25]. The TTA created predictions based on the original spectrogram from our dataset and four augmentations of it. The mean prognosis from all the images was equal to 77%.

The proposed method produced a promising classification accuracy of around 70% for a ResNet-34 model, and 71% for a ResNet-50 model, both trained on spectrograms of 224x224 px. This result can be improved to 77% with TTA. A full summary of the results is presented in Table 2. The variations in the results for each system configuration is attributable to the fact that both the training and test datasets were randomly selected at each system initialization stage.

---

[‡‡‡‡]Link to Google Colab notebook: https://colab.research.google.com/drive/1MHcM2uBgY_MgjwAJbz_TSjlLXBdaJgpm
[§§§§]Fast.ai uses PyTorch and provides a single consistent API to the most important deep learning applications and data types. More information: https://www.fast.ai/

Table 2. Summary of other classification results for different CNN architectures.

| Model | Type of Input | Hyperparameters (LR; EP) | Accuracy | F1 Score | Precision | Recall |
|---|---|---|---|---|---|---|
| ResNet 18 | Image (224x224px) | 0.01; 3 | 78% | 0.0000 | - | 0.0000 |
| ResNet 34 | Image (224x224px) | 0.001; 3 | 81% | 0.6154 | 0.5714 | 0.6667 |
| ResNet 50 | Image (224x224px) | 0.01; 3 | 67% | 0.3077 | 0.3333 | 0.2857 |
| ResNet 50 | Image (512x512px) | 0.01; 3 | 78% | 0.5714 | 0.5714 | 0.5714 |
| ResNet 101 | Image (512x512px) | 0.001; 4 | 81% | 0.2857 | 0.2500 | 0.3333 |
| ResNet 101 | Image (1024x1024px) | 0.0044; 4 | 78% | 0.4000 | 0.2500 | 1.0000 |
| ResNet 152 | Image (512x512px) | 0.0001; 3 | 63% | 0.3750 | 0.3750 | 0.3750 |
| ResNet 152 | Image (1024x1024px) | 0.003; 3 | 74% | 0.5333 | 0.5000 | 0.5714 |

The ResNet-34 system using a smaller Set A and TTA classified 21 voice samples correctly, with only six samples classified incorrectly. The ResNet-50 system using a larger Set B classified 443 voice samples correctly, with only 199 samples classified incorrectly. The confusion matrixes for both configurations are presented in Fig. 8.

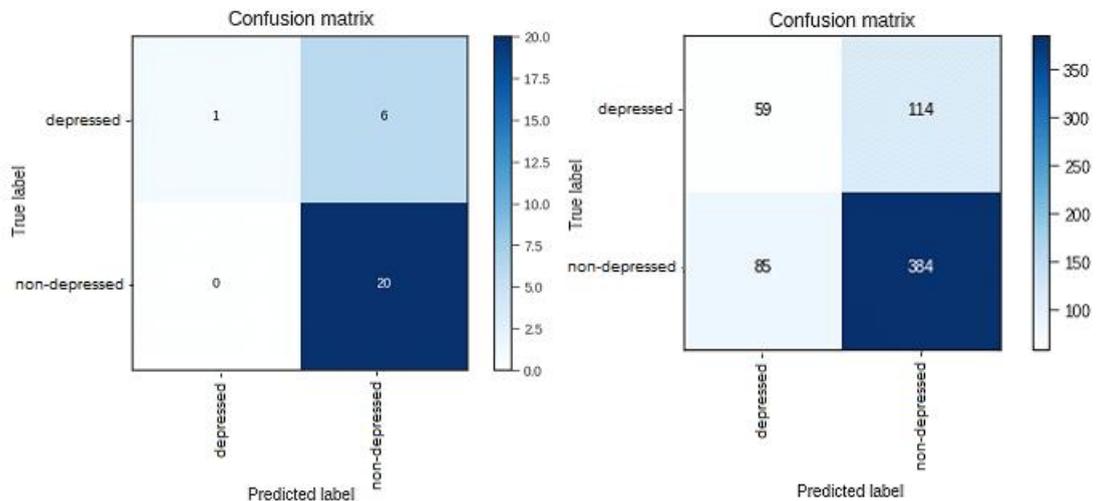

Fig. 8. Confusion Matrix for our classification system on (a) Set A and (b) Set B using images of 224x224px.

## 5. Conclusions and future work

This paper proposed a method that uses deep convolutional neural networks for depression detection in speech. We tested five network architectures and selected ResNet-34 and ResNet-50 as presenting the best classification results. The results suggest a promising new direction in using audio spectrograms for preliminary screening of depressive subjects by using short samples of their voice. The spectrograms proved to have a potential for generating CNN learnable features. The algorithm attained accuracies of 70% and 77% using the TTA method. This was despite the challenging nature of voice as a predictor of depression. The solution used small sample sizes (15 seconds), which diminished the effect of noise, in our view.

Our system can be used independently, or as an element of a more complex, hybrid, or multimodal solution. The main advantages of this method are its relative simplicity, coupled with its state-of-the art accuracy. The solution achieved accuracy similar to Afshan et al. [12] on short speech utterances. The classification sensitivity (recall) of 41% is 3% higher than a context-free model [14] and achieves the same precision as a weighted (audio) model [14].

In future work, we will evaluate the performance of our method for a different set of audio sample sizes (10, 20, 30, 40 seconds, and 1.8 minutes) to ensure better comparability of results with other researchers from the Interspeech conference [12]. In addition, there are more advanced options to convert audio recordings into images that could improve the classification quality further by extracting more features from the same size audio samples. As mentioned above, there is also the option of using our method as a building block for a more complex, hybrid, or multimodal depression detection system that could focus on PHQ-8 score prediction. We will also conduct all future experiments on a fixed, balanced test and training dataset.

Further training and validation will be performed on the new dataset currently being gathered by the Interdisciplinary Centre for Applied Cognitive Studies (ICACS) at SWPS University as a part of the Horizon 2020 funded project [27]. The data being gathered contain participant audio files, questionnaires, and eye-tracking points; thus, the potential exists to perform additional research.

We believe that the proposed system can be used as an innovative depression screening system and can help both the community of clinical therapists and patients.

## References

[1] World Health Organization (2017) "WHO global health days - Staying positive and preventing depression as you get older." Retrieved from https://www.who.int/campaigns/world-health-day/2017/handouts-depression/older-age/en/
[2] World Health Organization. (2018) "Agenda Item 12.4 Digital Health." *In Proceedings of Seventy-First World Health Assembly*, 21–26 May 2018 in Geneva, Switzerland. (pp. 2-3). http://apps.who.int/gb/ebwha/pdf_files/WHA71/A71_R7-en.pdf
[3] Quatieri, Thomas F., and Nicolas Malyska (2012) "Vocal-source biomarkers for depression: A link to psychomotor activity." *In Thirteenth Annual Conference of the International Speech Communication Association*.
[4] Cummins, Nicholas, Stefan Scherer, Jarek Krajewski, Sebastian Schnieder, Julien Epps, and Thomas F. Quatieri (2015) "A review of depression and suicide risk assessment using speech analysis." *Speech Communication*, **71**: 10-49.
[5] Ringeval, Fabien, Björn Schuller, Michel Valstar, Jonathan Gratch, Roddy Cowie, Stefan Scherer, Sharon Mozgai, Nicholas Cummins, Maximilian Schmitt, and Maja Pantic (2017, October) "Avec 2017: Real-life depression, and affect recognition workshop and challenge." *In Proceedings of the 7th Annual Workshop on Audio/Visual Emotion Challenge* (pp. 3-9). ACM.
[6] Yang, Le, Hichem Sahli, Xiaohan Xia, Ercheng Pei, Meshia Cédric Oveneke, and Dongmei Jiang (2017, October) "Hybrid depression classification and estimation from audio video and text information." *In Proceedings of the 7th Annual Workshop on Audio/Visual Emotion Challenge* (pp. 45-51). ACM.
[7] Kroenke, Kurt, Tara W. Strine, Robert L. Spitzer, Janet BW Williams, Joyce T. Berry, and Ali H. Mokdad (2009) "The PHQ-8 as a measure of current depression in the general population." *Journal of affective disorders*, **114** (1-3): 163-173.
[8] Willmott, Cort J., and Kenji Matsuura (2005) "Advantages of the mean absolute error (MAE) over the root mean square error (RMSE) in assessing average model performance." *Climate research*, **30** (1): 79-82.
[9] Kotsiantis, Sotiris B., I. Zaharakis, and P. Pintelas (2007) "Supervised machine learning: A review of classification techniques." *Emerging artificial intelligence applications in computer engineering*, **160**: 3-24.
[10] Yang, Le, Dongmei Jiang, Xiaohan Xia, Ercheng Pei, Meshia Cédric Oveneke, and Hichem Sahli (2017, October) "Multimodal measurement of depression using deep learning models." *In Proceedings of the 7th Annual Workshop on Audio/Visual Emotion Challenge* (pp. 53-59). ACM.
[11] Shin, Hoo-Chang, Holger R. Roth, Mingchen Gao, Le Lu, Ziyue Xu, Isabella Nogues, Jianhua Yao, Daniel Mollura, and Ronald M. Summers (2016) "Deep convolutional neural networks for computer-aided detection: CNN architectures, dataset characteristics and transfer learning." *IEEE transactions on medical imaging*, **35** (5): 1285-1298.
[12] Afshan, Amber, Jinxi Guo, Soo Jin Park, Vijay Ravi, Jonathan Flint, and Abeer Alwan (2018) "Effectiveness of voice quality features in detecting depression." *In Proc. Interspeech* (pp. 1676-1680).
[13] Molla, K. I., and Keikichi Hirose (2004, July) "On the effectiveness of MFCCs and their statistical distribution properties in speaker identification." *In 2004 IEEE Symposium on Virtual Environments, Human-Computer Interfaces and Measurement Systems, 2004.(VCIMS).* (pp. 136-141). IEEE.
[14] Al Hanai, Tuka, Mohammad Ghassemi, and James Glass (2018) "Detecting depression with audio/text sequence modeling of interviews." *In Proc. Interspeech* (pp. 1716-1720).
[15] Gratch, Jonathan, Ron Artstein, Gale Lucas, Giota Stratou, Stefan Scherer, Angela Nazarian, Rachel Wood, Jill Boberg, David DeVault, Stacy Marsella, David Traum, Albert Rizzo, and Louis-Philippe Morency (2014) "The distress analysis interview corpus of human and computer interviews." *In Proceedings of the Ninth International Conference on Language Resources and Evaluation*, LREC. ELRA, Reykjavik, Iceland, 3123–3128.
[16] DeVault, David, Ron Artstein, Grace Benn, Teresa Dey, Ed Fast, Alesia Gainer, Kallirroi Georgila et al. (2014) "SimSensei kiosk: A virtual human interviewer for healthcare decision support." *In Proceedings of the 13th International Conference on Autonomous Agents and Multiagent Systems (AAMAS'14)*, Paris
[17] Tzanetakis, George, Georg Essl, and Perry Cook (2001, September) "Audio analysis using the discrete wavelet transform." *In Proc. Conf. in Acoustics and Music Theory Applications* (Vol. 66).

[18] Le, Bin, Thomas W. Rondeau, Jeffrey H. Reed, and Charles W. Bostian (2005) "Analog-to-digital converters." *IEEE Signal Processing Magazine*, **22** (6): 69-77.
[19] Pytorch (2018) "Torchvision models." https://pytorch.org/docs/stable/torchvision/models.html
[20] He, Kaiming, Xiangyu Zhang, Shaoqing Ren, and Jian Sun (2016) "Deep residual learning for image recognition." *In Proceedings of the IEEE conference on computer vision and pattern recognition* (pp. 770-778).
[21] Zawadzka-Gosk, Emilia, Krzysztof Wołk, and Wojciech Czarnowsk (2019, April) "Deep learning in state-of-the-art image classification exceeding 99% accuracy." In *World Conference on Information Systems and Technologies* (pp. 946-957). Springer, Cham.
[22] Johnson, Rie, and Tong Zhang (2013) "Accelerating stochastic gradient descent using predictive variance reduction." *Advances in neural information processing systems* 315-323.
[23] Kingma, Diederik P., and Jimmy Ba (2014) "A method for stochastic optimization." *arXiv preprint arXiv*:1412.6980
[24] Tharwat, Alaa (2018) "Classification assessment methods." *Applied Computing and Informatics*.
[25] Brownlee, Jason (2019) "How to use test-time augmentation to improve model performance for image classification." Retrieved from https://machinelearningmastery.com/how-to-use-test-time-augmentation-to-improve-model-performance-for-image-classification/
[26] Huang, Gao, Yixuan Li, Geoff Pleiss, Zhuang Liu, John E. Hopcroft, and Kilian Q. Weinberger (2017) "Snapshot ensembles: Train 1, get m for free." *arXiv preprint arXiv*:1704.00109.
[27] SWPS University of Social Sciences and Humanities (2018) "Dynamic Model of Recurring Negative Thoughts – Method of Daily Measurement" Retrieved from https://www.swps.pl/nauka-i-badania/granty/16335-dynamiczny-model-powtarzajacych-sie-mysli-negatywnych-i-hamowania-w-depresji-metoda-codziennych-pomiaro
[28] Sak, Haşim, Andrew Senior, and Françoise Beaufays (2014) "Long short-term memory recurrent neural network architectures for large scale acoustic modeling." *In Fifteenth annual conference of the international speech communication association*.